\pdfoutput=1
\documentclass[11pt]{article}

\usepackage[]{ACL2023}

\usepackage{times}
\usepackage{latexsym}
\usepackage{amsmath}
\usepackage{acronym}
\usepackage{cprotect}
\usepackage{lipsum}% http://ctan.org/pkg/lipsum
\usepackage{graphicx}% http://ctan.org/pkg/graphicx
\usepackage{multirow}
\usepackage{hyperref}

\usepackage[norelsize, linesnumbered, ruled, lined, boxed, commentsnumbered]{algorithm2e}

\usepackage[T1]{fontenc}
\usepackage[utf8]{inputenc}

\usepackage{microtype}
\usepackage{enumitem}
\setlist{noitemsep,topsep=0pt,leftmargin=10pt}

\SetKwComment{Comment}{/* }{ */}
\title{Simple yet Effective Code-Switching Language Identification\\with Multitask Pre-Training and Transfer Learning}

\author{Shuyue Stella Li, Cihan Xiao, Tianjian Li, Bismarck Odoom \\
         Center for Language and Speech Processing, Johns Hopkins University \\ 
         \texttt{sli136, cxiao7, tli104, bodoom1@jhu.edu}}

\begin{document}
\maketitle
\newacro{cmi}[CMI]{Code-Mixing Index}
\newacro{scm}[SCM]{Synthetic Code-Mixing}
\newacro{ncm}[NCM]{Natural Code-Mixing}
\newacro{pos}[POS]{Part-of-speech}
\newacro{nlp}[NLP]{natural language processing}
\newacro{sa}[SA]{Sentiment analysis}
\newacro{plm}[PLM]{Pre-trained Language Models}
\newacro{sra}[SRA]{Syntactic Replacement Algorithm}
\newacro{lra}[LRA]{Lexical Replacement Algorithm}
\begin{abstract}
Code-switching, also called code-mixing, is the linguistics phenomenon where in casual settings, multilingual speakers mix words from different languages in one utterance. Due to its spontaneous nature, code-switching is extremely low-resource, which makes it a challenging problem for language and speech processing tasks. In such contexts, Code-Switching Language Identification (CSLID) becomes a difficult but necessary task if we want to maximally leverage existing monolingual tools for other tasks. In this work, we propose two novel approaches toward improving language identification accuracy on an English-Mandarin child-directed speech dataset. Our methods include a stacked Residual CNN+GRU model and a multitask pre-training approach to use Automatic Speech Recognition (ASR) as an auxiliary task for CSLID.
Due to the low-resource nature of code-switching, we also employ careful silver data creation using monolingual corpora in both languages and up-sampling as data augmentation. 
We focus on English-Mandarin code-switched data, but our method works on any language pair. Our best model achieves a balanced accuracy of 0.781 on a real English-Mandarin code-switching child-directed speech corpus and outperforms the previous baseline by 55.3\%.
\end{abstract}

\section{Introduction}
With more than 6000 languages still alive today, there are more people speaking more than one language (whether from birth or through late acquisition) than monolingual speakers \cite{marian2012cognitive}. When multilingual speakers who share two or more of the same languages engage in a conversation, they naturally tend to switch languages spontaneously. Code-switching allows bilingual speakers to express their intentions more freely and to be better understood \cite{heredia2001bilingual}.
With the development of machine learning and neural networks, language and speech processing with most high-resource monolingual languages are highly effective. However, it is non-trivial to adapt the monolingual tools to multilingual and code-switching tasks. Additionally, as demonstrated in our later experiments, even large models trained on multilingual data such as Whisper \cite{radford2022whisper} and XLSR \cite{babu2021xls, conneau2020unsupervised} are limited in processing code-switched data between two high-resource languages. Therefore, an effective approach to leverage existing monolingual or multilingual pre-trained speech and language models and other NLP tools is to identify the language in each segment of a code-switched speech or text. 

\begin{figure}[t]
\centering
\includegraphics[width=0.4\textwidth]{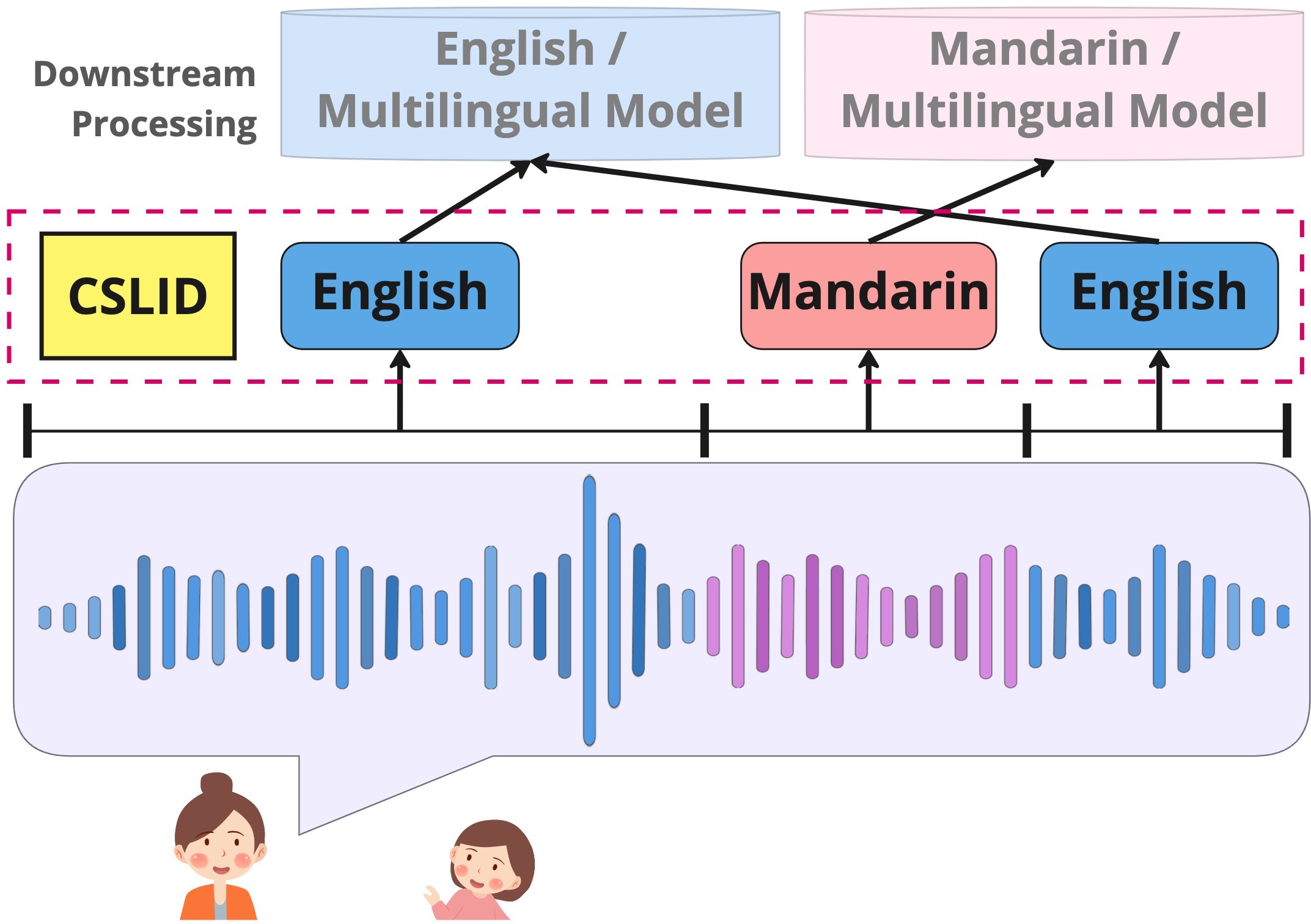}
\caption{``Split-then-Process'' Pipeline. The red dotted box is the main focus of our study. Given a segmented utterance from a child-directed domain, the language of each segment is identified by our system. This can potentially be useful for a range of downstream processing tasks that leverages existing monolingual or multilingual tools.}\label{fig:story}\vspace{-4mm}
\end{figure}

In this work, we focus on the language identification task of code-switched English and Mandarin speech data collected from Singapore on a child-directed activity. Singapore is such a language-dense region where four primary languages are spoken by the people - English, Malay, Mandarin, and Tamil, and almost all Singaporeans are bilingual or multilingual. The language diversity in the region contributes to the wide dialectal variations of the code-switched data and increases the difficulty of speech-processing tasks. The child-directed characteristic of the data makes the problem unique in that both the content domain and the speech style deviate from standard datasets and models. Domain mismatch problems have been addressed by data augmentation \cite{sun2021domain} or unsupervised adversarial training \cite{wang2018unsupervised} approaches in Automatic Speech Recognition (ASR) and gradual fine-tuning (GFT) in text-based settings \cite{xu2021gradual}. We adopt both data augmentation and GFT to the speech CSLID task in this work to improve the robustness of our system.

As illustrated in Figure \ref{fig:story}, the main objective of our model is to identify the language of a segment of speech, so that monolingual or multilingual models can be more effectively used for downstream tasks. With English and Mandarin being the languages with the largest number of speakers in Singapore and high-resource languages in the world, Code-Switching Language Identification is a crucial step in the ``Split-then-Process'' pipeline. Possible downstream tasks that could benefit from a robust language identification system include speech recognition, speech synthesis, or speech translation. We leverage monolingual data such as AISHELL (Mandarin) \cite{aishell_data} and LibriSpeech (English) \cite{panayotov2015librispeech} to build a CSLID model that is robust to both domain and dialectal variations.
The main contributions of our work are summarized as follows:
\begin{itemize}
    \item We propose two systems for code-switching language identification, a Residual CNN with BiRNN network (CRNN) and an Attention-based Multitask Training Model with combined ASR and CSLID loss. The systems can be easily extended to any language pair.
    \item We investigate the effect of pre-training with data augmentation from monolingual sources and the effect of fine-tuning with out-of-domain code-switched data, concluding that data balance is more crucial than domain similarity.
    \item We demonstrate that small and efficient architectures with effective data augmentation can be extremely successful in the CSLID task, outperforming massive multilingual pre-trained language models (PLM). Our system placed 2nd in a challenge featuring an English-Mandarin code-switching child-directed speech corpus [reference redacted for review], and we make our code publicly available\footnote{We make the project open source at [link hidden for review].} for further explorations in the field of code-switching speech processing.
\end{itemize}

\section{Related work}
Due to the increase of globalization and the growing population of bilingual and multilingual speakers, there is an emerging need for better language technologies for code-switching languages. Due to its spontaneous nature, code-switching happens more in colloquial settings, making it difficult for data collection. Code-switching is also a complex sociocultural linguistic phenomenon that depends on a combination of factors including topic, formality, and speaker intent \cite{mabule2015code, nilep2006code}. Code-switching can happen at different levels of the utterance (intersentential, intrasentential, intra-word) \cite{myers1989codeswitching}.
%In a single utterance, speakers can use different languages for each sentence (inter-sentential), switch halfway through a sentence (intra-sentential), insert words or phrases (emblematic), or use the syntactic rules of one language on the other, such as inflections (intra-word) \cite{myers1989codeswitching}. 
% Code-switching also happens in a diverse range of settings, including native bilingual conversations \cite{reyes2004functions}, language acquisition \cite{fanani2018code, martin1995code}, informal interviews \cite{yuliana2015code}, and even novels from multilingual communities \cite{adi2018code}.
All the above characteristics make code-switching a fascinatingly diverse and challenging topic of study.
In both text and speech processing, CSLID is a crucial step for downstream tasks such as text normalization for text-to-speech synthesis \cite{manghat2022normalization}, part-of-speech tagging \cite{solorio2008part}, speech translation \cite{weller2022end}, and speech recognition \cite{zhang2021rnn, zhang2022streaming, zhou2022configurable, sreeram2020exploration}.

\subsection{Multidialectal Code-Switching}
Code-switching speech processing faces the issue of dialectal variations.%, especially in a region like Singapore, where a variety of dialects exist for each language spoken in the region. 
In Singapore, Mandarin, Hokkien, and Cantonese are the major Chinese dialects with most speakers, along with Teochew, Hakka, and Hainanese \cite{gupta1995language}. 
%With the bilingual education policy in Singapore, code-switching becomes extremely common in the Chinese-ethnic community \cite{hornberger2009multilingual}. 
\cite{chowdhury2021towards} proposed an end-to-end attention-based conformer architecture for multi-dialectal Arabic ASR. \cite{rivera2019automatic} built an acoustic model for code-switching detection among Arabic dialects. However, there is a lack of sufficient research on code-mixing between non-standard Mandarin and non-standard English, which is the focus of our study.

\subsection{Code-Switching Language Identification}
The use of Convolutional Neural Networks (CNN) in speech processing is widely adopted due to the use of spectrogram or filter bank as the first feature extraction step of speech signal processing in monolingual tasks \cite{ganapathy2014robust}. Deep Neural Networks (DNN) \cite{yılmaz2016dnn} and phoneme units-based Hidden Markov Model (HMM) with Support Vector Machine (SVM) classifier \cite{mabokela2014modeling, mabokela2013integrated} have also been used for CSLID. Additionally, CSLID is often integrated into ASR systems as an auxiliary task to improve the ASR performance \cite{lounnas2020cliasr, shan2019investigating}. However, these approaches have a different focus from our current study, which aims to improve the CSLID performance for a range of speech-processing tasks.

\subsection{Data Augmentation \& Multilingual PLMs}
Various data augmentation techniques have been used for code-switching, but mostly focused on text processing tasks \cite{xu2021can, li2022language}. Some work uses text-based data augmentation for speech tasks \cite{hussein2023textual, nakayama2019recognition}. \cite{ali2021arabic} uses monolingual English and Arabic speech data for the code-switched ASR task. However, there is little prior work to synthetically generate code-switched speech data from monolingual sources. In our work, we segment monolingual speech data in the sub-utterance level to simulate code-switched speech data augmentation.

% 
% \subsection{Multilingual PLMs}
Additionally, with the recent development of massively multilingual pre-trained speech and language models such as mSLAM \cite{bapna2022mslam}, Whisper \cite{radford2022whisper} and XLS-R \cite{babu2021xls, conneau2020unsupervised}, it is easier to leverage monolingual data for multilingual tasks. The use of multilingual PLMs for code-switching tasks in the text domain has proven to be successful \cite{rathnayake2022adapter}, but it has not been widely used in the speech setting due to limited data and costly training. In our work, we use the multilingual PLMs as a zero-shot baseline with which we compare our parameter-efficient models.
%https://arxiv.org/pdf/2212.09660.pdf

\section{Methodology}

% \begin{figure*}[ht]
% \centering
% \includegraphics[width=\textwidth,height=49mm]{figures/systems.jpeg}
% \caption{Proposed Systems}\label{fig:systems}
% \end{figure*}

In this section, we first propose three systems for the CSLID task and describe the architectural design tailored to the different characteristics of the data and model. Then, we introduce our data augmentation method leveraging out-of-domain code-switching data with a GFT schedule to improve upon the pre-train-fine-tune paradigm.

\subsection{CRNN}
\begin{figure}[ht]
\centering\vspace{-1mm}
\includegraphics[width=0.48\textwidth]{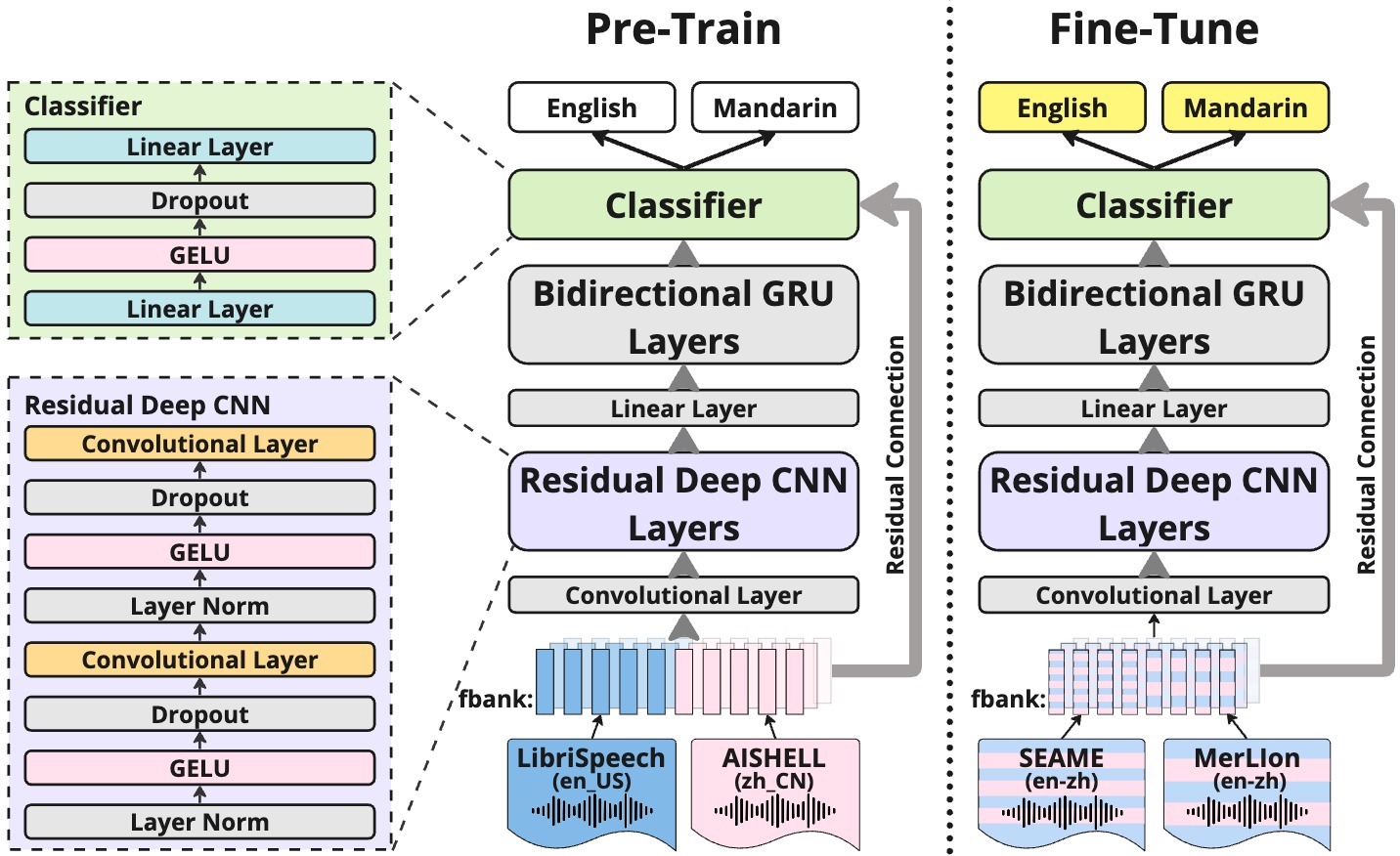}
\caption{CRNN Model}\label{fig:crnn}
\end{figure}
The CRNN model, inspired by \cite{crnn-lid}, is a stack of Residual CNNs and RNNs. We utilize the power of CNN to extract features directly from the spectral domain and use an RNN to extract temporal dependencies. We use bi-GRU \cite{cho-etal-2014-learning} layers for the RNN component of the model because it has fewer parameters, making it faster to train and less prone to over-fitting.
As illustrated in Figure \ref{fig:crnn}(a), our CRNN model is a simplified version of \cite{crnn-lid} with 3 CNN layers to extract acoustic features and 5 GRU layers with hidden dimension 512 to learn features for language identification. We apply a linear classifier to the last hidden state of the RNN.  

\begin{table*}[ht]
\centering
\scalebox{0.9}{\begin{tabular}{c|cccc|cccc|ccc|cc}
\hline
\multirow{2}{*}{\textbf{Stage}}   & \multicolumn{4}{c|}{\textbf{MERLion (M)}} & \multicolumn{4}{c|}{\textbf{SEAME (S)}} & \multicolumn{3}{c|}{\textbf{Total}} & \multicolumn{2}{c}{\textbf{Ratios}}\\\cline{2-14}
& zh & en & total & zh/en & zh & en & total & zh/en & zh & en & total  & S/M & zh/en\\\hline
% 1 & \hspace{2mm}5.4 (1) & 21.6 & 27.0 & 0.2 & 48.3 & 32.2 & 80.4 & 1.5 & 53.6 & 53.8 & 107.4 & 3.0 & 1.0 \\
1 & \hspace{2mm}5.4 (1) & 21.6 & 27.0 & 0.2 & 17.9 & 8.9 & 26.8 & 2.0 & 23.2 & 30.6 & 53.8 & 1.0 & 0.76 \\
2 & 10.7 (2)            & 21.6 & 32.4 & 0.5 & 10.7 & 5.4 & 16.1 & 2.0 & 21.4 & 27.0 & 48.4 & 0.5 & 0.79 \\
3 & 10.7 (2)            & 21.6 & 32.4 & 0.5 & 4.5  & 2.2 & 6.7  & 2.0 & 15.2 & 23.9 & 39.1 & 0.2 & 0.64 \\
4 & 16.1 (3)            & 21.6 & 37.7 & 0.7 & 0.0  & 0.0 & 0.0  & -   & 16.1 & 21.6 & 37.7 & 0.0 & 0.74 \\ \hline
\end{tabular}}
\caption{\label{tab:gradual} Grdual FT Schedule. Values inside parentheses are up-sampling ratios for the MERLion zh utterances.}
\end{table*} 

\subsection{Multi-Task Learning (MTL)}
\begin{figure}[ht]
\centering\vspace{-1mm}
\includegraphics[width=0.48\textwidth]{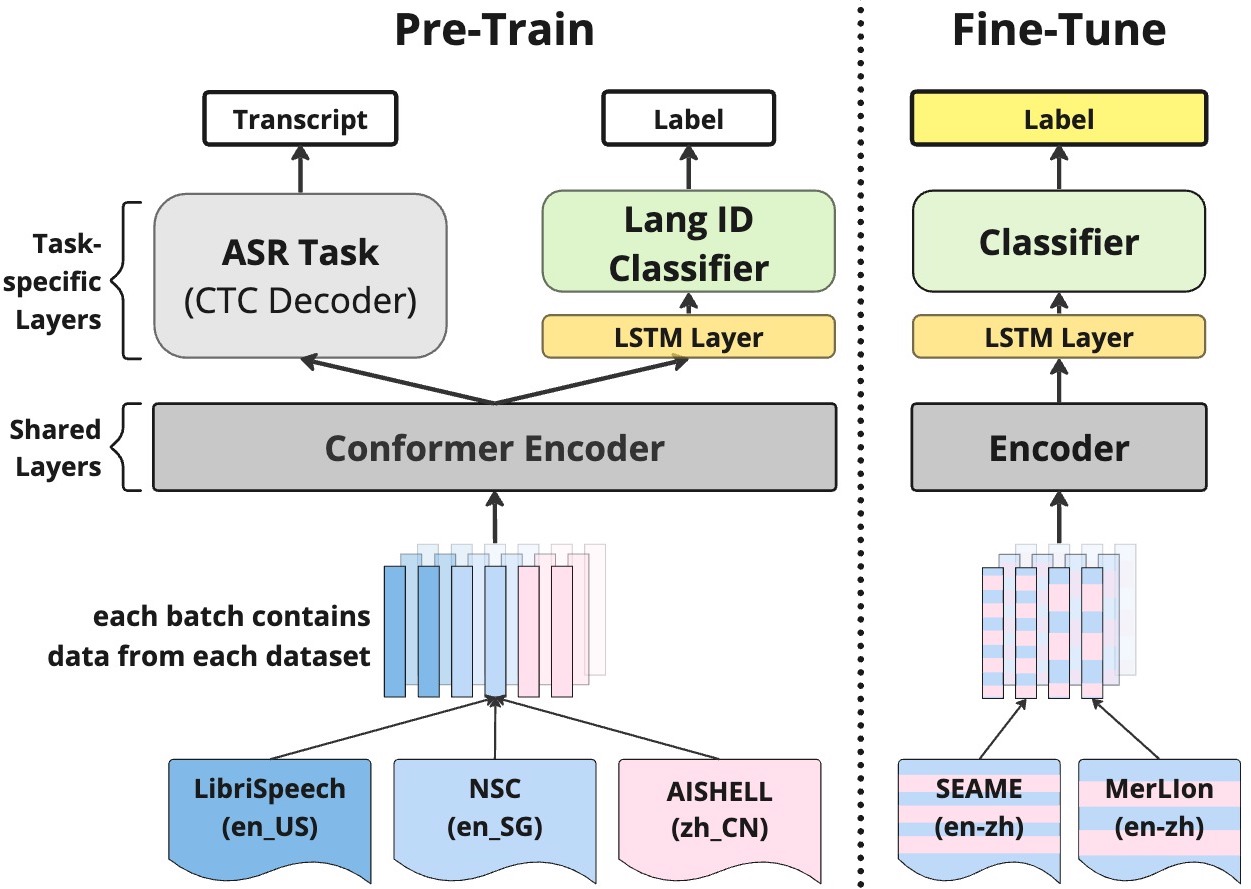}
\caption{Multi-task Learning Model}\label{fig:mtl}
\end{figure}
To enhance the model's ability to extract acoustic features, we train a model via multitask learning with a joint Connectionist Temporal Classification (CTC) and LID loss, as illustrated in Figure~\ref{fig:mtl}(b). The architecture of the model is based on a Conformer encoder, along with a linear layer for CTC decoding and an LSTM + linear layer for LID decoding. Similar to the CRNN model, we conduct phased training to first pre-train the Conformer model with the joint loss on the monolingual corpora and fine-tune the model on the MERLIon and SEAME datasets with only the LID loss. This approach aims to better adapt the model to the target classification task.

\subsection{Multilingual PLMs}
Being pre-trained on multiple languages, massively multilingual PLMs are a powerful tool for cross-lingual tasks. We want to understand the out-of-the-box ability of PLMs to process code-switching sentences by comparing the zero-shot CSLID performance of Whisper \cite{radford2022whisper} and XLSR \cite{babu2021xls, conneau2020unsupervised} against the more parameter-efficient models we introduce in this work. For Whisper, we use the \verb+detect_language()+ method from the model class, passing in CutSets with a max duration of 50. For XLSR, we perform  two-way zero-shot classification using \url{wav2vec2-xls-r-300m} with a LID head. The LID head is a 2-layer Bidirectional GRU with a linear layer.

\subsection{Data Augmentation}\label{sec:gradual}
Child-directed English-Mandarin code-switching is an extremely low resource problem. As such, we propose a data augmentation method that takes advantage of any additional data in a similar distribution to improve the performance of the model. The target in domain data - MERLion - is unbalanced such that the ratio of English to Mandarin labels is about 4:1. In addition to up-sampling the Mandarin utterances during training, our proposed data augmentation approach combines the SEAME code-switching dataset (as described in detail in Section \ref{sec:dataset}) that has more Mandarin utterances than English ones. Lastly, we propose a gradual fine-tuning schedule for smooth domain adaptation as described in Table \ref{tab:gradual} below \cite{xu2021gradual}. As we up-sample the Mandarin utterances in the MERLion dataset and vary the ratio of Mandarin to English in the sampled SEAME dataset to control for a smooth transition to the real Mandarin-English ratios in the development set. The gradual FT terminates with a stage of using only the MERLion dataset (with Mnadarin up-sampled) without the out-of-domain SEAME data. All the experiments described in Table \ref{tab:gradual} are fine-tuning the model checkpoint pre-trained on monolingual Mandarin and English speech.

\section{Experiments}

\subsection{Dataset \& Metric}\label{sec:dataset}
We use multiple monolingual English and Mandarin and code-switched English-Mandarin datasets in our experiments, including LibriSpeech \cite{panayotov2015librispeech}, National Speech Corpus of Singapore (NSC) \cite{NSC_data}, AISHELL \cite{aishell_data}, SEAME \cite{SEAME_data}, and MERLIon \cite{chua2023development}. Table \ref{tab:data} reports the language and size of each dataset. Note that not all datasets are used for each experiment. The MERLIon dataset is split into training and development sets, and we refer to the train split of the MERLIon dataset as ``MERLIon'' in our system descriptions.

\begin{table}[ht]
\centering
\begin{tabular}{@{}l|cc@{}}
\hline
Dataset   & Language & Length (hr) \\ 
\hline
LibriSpeech-clean  & en (US)   & 100    \\
NSC & en (SG) & 100   \\
AISHELL   &  zh    &  200      \\
SEAME    & en-zh & 100  \\
MERLIon &  en-zh   &  30 \\
\hline
\end{tabular}\vspace{-1mm}
\caption{Datasets used in our experiments.}
\label{tab:data}\vspace{-3mm}
\end{table}

\paragraph{Metric}The MERLIon dataset roughly contains 25 hours of English speech and 5 hours of Mandarin speech. Due to this severe data imbalance issue, we use the Balance Accuracy (BAC), which is the average of recall obtained for each label class, rather than the Absolute Accuracy as the metric to evaluate our systems. In the submission of the English-Mandarin code-switching task, the evaluation also reports the Equal Error Rate (EER), which is defined to be the threshold for an equal false acceptance rate and false rejection rate.

\paragraph{Baseline} The baseline over which we attempt to improve is the system developed by the task organizers, which consists of an end-to-end conformer model trained on the same available data \cite{chua2023development}. This system has a BAC of 50.32\% and an EER of 22.13\%.

\subsection{Preprocessing}
We preprocess the data using lhotse\footnote{https://github.com/lhotse-speech/lhotse}, a Python toolkit designed for speech and audio data preparation. We standardize the sample rate of all audio recordings to 16kHz by downsampling utterances in the development and test dataset with sample rates $>$ 16kHz. Prior to training, we extract 80-dimensional filterbank (fbank) features from the speech recordings and apply speed perturbation with factors of 0.9 and 1.1. During training, we use on-the-fly SpecAug~\cite{specaug} augmentation on the extracted filter bank features with a time-warping factor of 80.

% Write about text tokenization and romanization
To train the model jointly with an ASR CTC loss, we first tokenize and romanize the bilingual transcripts with space-delimited word-level tokenization for monolingual English transcripts (LibriSpeech and NSC) and monolingual Mandarin transcripts in AISHELL, as these transcripts were pre-tokenized and separated by spaces. For the occasionally code-switched Mandarin words in NSC, we remove the special tags and kept only the content of the Mandarin words. The SEAME dataset contains a portion of untokenized Mandarin transcripts. Hence, we tokenize all Mandarin text sequences with length $>4$ using a Mandarin word segmentation tool jieba\footnote{https://github.com/fxsjy/jieba}. Additionally, to reduce the size of the model, we adopt a pronunciation lexicon, utilizing the CMU dictionary for English word-to-phoneme conversion and the python-pinyin-jyutping-sentence tool for generating the pinyin for Mandarin words\footnote{https://github.com/Language-Tools/python-pinyin-jyutping-sentence}. To enhance the model's ability to capture the lexical information in the training data, we add a suffix "\_cn" for Mandarin phonemes.

\subsection{Experimental Setup} 
We follow the pre-train-fine-tune paradigm for most experiments except for the zero-shot PLM baseline and ablation experiments to investigate the effect of pre-training. In the pre-training stage, we use the monolingual datasets (LibriSpeech, AISHELL, and NSC), and in the fine-tuning state, we use the code-switched datasets (SEAME and MERLIon). 

\begin{table}[h]
\centering
\scalebox{0.78}{\begin{tabular}{@{}r|c|c|c|c@{}}
\hline
\# & System & PT Data & FT Data & FT Method \\ 
\hline
1  & \multirow{7}{*}{CRNN}  &                              & -       & -       \\\cline{4-5}
2  &                        & LibriSpeech                  & MERLIon & 1-stage \\\cline{4-5}
3  &                        & + AISHELL                    & MERLIon & combined \\\cline{5-5}
4  &                        &                              & + SEAME & gradual \\\cline{3-5}
5  &                        & \multirow{3}{*}{ - }         & MERLIon & 1-stage \\\cline{4-5}
\multirow{2}{*}{6}&         &                              & MERLIon & \multirow{2}{*}{combined} \\
   &                        &                              & + SEAME &          \\\hline
7  & \multirow{2}{*}{MTL} & LibriSpeech     & MERLIon & combined \\\cline{5-5}
8  &                        & + AISHELL + NSC              & + SEAME & 2-stage \\\hline
9  & Whisper                & - & - & - \\
10  & XLSR                & - & - & - \\
\hline
\end{tabular}}\vspace{-1mm}
\caption{Experimental Setup for Basic Experiments}
\label{tab:exp}\vspace{-3mm}
\end{table}

Table \ref{tab:exp} shows the experiments conducted in our study along with the pre-training and fine-tuning datasets and fine-tuning method. For this set of experiments, FT Methods: 1-stage FT means fine-tuning the model on the MERLIon dataset only; combined FT is fine-tuning the model on a 1-1 proportion of SEAME and MERLIon data; and gradual FT is fine-tuning the model on more SEAME (out-of-domain) data than MERLIon (in-domain) data, then increasing the ratio of MERLIon data gradually until the fine-tuning set contains only MERLIon data.

Table \ref{tab:exp2} summarizes the second set of experiments involving the up-sampling with  schedule described in Section \ref{sec:gradual}. Note that in Experiment \#15, only the 1:1 mix of MERLion:SEAME is used as a control for the  setting for the MTL system.
\begin{table}[h]
\centering
\scalebox{0.95}{\begin{tabular}{@{}r|c|c|c|c@{}}
\hline
\# & System & epoch/stage & total epochs & LR \\ 
\hline
11  & \multirow{4}{*}{CRNN}  &  \multirow{2}{*}{3}  & \multirow{2}{*}{12} & 0.001 \\\cline{5-5}
12  &                        &                      &                   & 0.00001 \\\cline{3-5}
13  &                        &  \multirow{2}{*}{5}  & \multirow{2}{*}{20} & 0.001 \\\cline{5-5}
14  &                        &                      &                   & 0.00001 \\\hline
\# & System & epoch range & total epochs & LR \\ \hline
15  & \multirow{5}{*}{MTL}   &  1-20                & 20 & \multirow{5}{*}{0.00001} \\\cline{3-4}
16  &                        &  1-5                 & 5  &  \\\cline{3-4}
17  &                        &  5-10                & 10 &  \\\cline{3-4}
18  &                        &  10-15               & 15 &  \\\cline{3-4}
19  &                        &  15-20               & 20 &  \\\hline
\end{tabular}}
\caption{Up-Sampling Experiments.}
\label{tab:exp2}
\end{table}

\subsection{Training}

\subsubsection{CRNN Training}
We pre-train our CRNN model for 5 epochs on 100 hours of clean speech from LibriSpeech\cite{panayotov2015librispeech} and 200 hours of preselected partition from AISHELL\cite{aishell_data}. Each batch contains a balanced amount of English and Mandarin sub-utterance level speech utterances to simulate an artificial speech code-switching dataset. We select the pre-trained model checkpoint with the best performance on the entire MERLIon dataset. Then, the model is fine-tuned on the MERLIon dataset (exp \#2) or the MERLIon+SEAME dataset (exp \#3\&4) for 10 epochs, leaving out 1 hour of MERLIon data (1749 English utterances and 100 Mandarin utterances) for evaluation. 
% Since we believe that there is an accent mismatch between the SEAME dataset and the MERLION dataset, we decided not to use the SEAME dataset during pre-training and fine-tuning. We perform experiments ablating the effect of pre-training and fine-tuning with or without the SEAME dataset and report our experiment results at table \ref{table:crnn}.
During training, we set the max duration of each cut to 120ms; we use the Adam optimizer with a pre-training learning rate of 1e-4 and a fine-tuning learning rate of 1e-5, with a dropout of 0.1. In the  experiment, ratios between the out-of-domain and in-domain data are $[3,2,1,0.5,0]$ over 5 epochs.

\subsubsection{Multitask Pre-Training}
The conformer model is pre-trained with the joint CTC/LID loss for 5 epochs as well on the monolingual data, including LibriSpeech, AISHELL, and NSC. To balance the loss for each task, we interpolate the losses with a hyperparameter $\lambda$. Formally, the overall loss $L$ is computed as below:
\begin{equation}
    L = (1 - \lambda) L_{\text{CTC}} + \lambda L_{\text{LID}} \cdot {\alpha}
\end{equation}
where $L_{\text{CTC}}$ denotes the CTC loss, $L_{\text{LID}}$ denotes the LID loss and $\alpha$ is the scaler for the LID loss. We set $\lambda = 0.2$ and $\alpha = 100$. The model is then fine-tuned for 15 epochs on the mixed MERLIon and SEAME datasets. We intentionally balance the total duration of samples drawn from each dataset, which implicitly biases toward the development set as it contains fewer utterances, and our sampler terminates when it finishes an epoch on the smaller corpus.

\section{Results and Analysis}

\begin{table}[ht]
\centering
\scalebox{0.82}{\begin{tabular}{@{}clccc@{}}
\hline
\# & experiment   & English & Mandarin & Balanced \\ \hline
1 & PT                  &  0.649   & 0.650    & 0.650  \\
2 & PT + FT (M)         &  0.927   & 0.630    & 0.779  \\
3 & PT + FT (M+S)       &  0.965   & 0.370    & 0.667  \\
4 & PT + FT (gradual)   &  0.851   & \textbf{0.720}    & \textbf{0.785}  \\
5 & FT (M)              &  1.0     & 0.0      & 0.5    \\
6 & FT (M+S)            &  0.988   &  0.09    & 0.539  \\\hline
7 & MTL + combined FT   &  0.960   &  0.610   & \textbf{0.785} \\
8 & MTL + 2-stage FT    & 0.957    &  0.46    & 0.708  \\\hline
9 & Whisper Zero-Shot   & 0.821    &  0.502   & 0.662  \\
10 & XLSR Zero-Shot     & 0.198    &  0.0     & 0.099   \\
\hline
\end{tabular}}
\caption{English, Mandarin and Balanced Accuracy of our CRNN model on the held-out development set of MERLIon. Table keys: \textbf{PT} = only pre-training, \textbf{FT} = fine-tuned on the MERLIon train split, \textbf{w/ SEAME} = fine-tuned with mixed MERLIon train split and SEAME dataset, \textbf{MTL} = multitask learning model with pre-training and fine-tuning. (All rows without PT indicate that the model parameters are randomly initialized.)}
\label{table:crnn}
\end{table}

\subsection{CRNN Results}
Table \ref{table:crnn} shows the English, Mandarin, and BAC of our CRNN model on the held-out part of the MERLIon development set.
The best-performing model is the model initialized from the best pre-train checkpoint and gradually fine-tuned on the MERLIon and SEAME dataset (PT+FT). Without , it is more effective to \textit{only} fine-tune on the MERLIon in-domain dataset, implying that directly combining out-of-domain sources (SEAME) causes additional complexity for the model. The Mandarin accuracies for training on the MERLIon dataset from scratch with (exp\#6) or without (exp\#5) the SEAME dataset are both poor - 0.0 for the model fine-tuning only on MERLIon and 0.09 for the model fine-tuning on the MERLIon and SEAME datasets.

\subsection{Multitask Pre-Training}
Two fine-tuning approaches were used for the MTL model. We find that after fine-tuning on the combined MERLIon + SEAME dataset, a second stage fine-tuning on only the MERLIon dataset in fact \textit{hurts} the performance. 
This might result from the imbalanced labeling effect, biasing the model toward the English predictions. Therefore, introducing more Mandarin samples from the SEAME corpus balances the labeling and yields better performance on the held-out set.

\subsection{Multilingual PLMs}
As shown in Table \ref{table:crnn}, the zero-shot performance of Whisper is not great but reasonable given the massive amount of data it was pre-trained on. However, zero-shot XLSR is extremely ineffective in doing CSLID. These results suggest that multilingual PLMs do not have the out-of-the-box capability to understand the complex phenomenon of code-switching and thus require careful fine-tuning. We report the performance of our CRNN model at task submission time on the MERLIon test set (labels unavailable to participants) in Table \ref{tab:results}.
%We notice that if the CRNN model is trained from scratch (no PT), it fails to learn anything and classifies all utterances as English when trained directly on the MERLIon dataset due to the heavy data imbalance issue. 

\begin{table}[h]
\centering
\scalebox{0.79}{\begin{tabular}{c|cc|cc}
\hline
\multirow{2}{*}{System} & \multicolumn{2}{c|}{Dev} & \multicolumn{2}{c}{Test} \\
       & EER & BAC & EER & BAC \\
\hline
 Baseline \cite{chua2023development} & - & - & 0.221 & 0.503 \\
 Whisper Zero-Shot & 0.228 & 0.662 & 0.230 & 0.649\\
 CRNN PT+FT & \textbf{0.146} & \textbf{0.663}  & \textbf{0.155} & \textbf{0.701} \\
\hline
\end{tabular}}
\caption{Equal Error Rate (EER) and Balanced Accuracy (BAC) on the MERLIon development and test sets for our submitted system and the previous baseline.}
\label{tab:results}\vspace{-3mm}
\end{table} 

\subsection{Ablation Studies} 
\subsubsection{Effect of Pre-Training}
As shown in exp \#5 and \#6, removing the pre-training stage results in significant performance drops. The model trained with only the MERLIon dataset classifies all utterances as English because the MERLIon dataset is heavily unbalanced, which contains 40287 English utterances and only 9903 mandarin utterances. 
This implies that the pre-training on monolingual data with balanced labels makes the model robust under heavily unbalanced classes, allowing the model to extract meaningful features for both languages even if data for one language is scarce. %We ablate the effect of pre-training by training on the MERLIon dataset from randomly initialized parameters (FT). We can see in table \ref{table:crnn} the performance heavily drops when the model is randomly initialized compared to the model initialized from the best pre-train checkpoint (PT + FT). 

\subsubsection{Effect of Code-Switched Fine-Tuning}
Directly using the pre-trained model (exp \#1) suffers from domain mismatch, suggesting that fine-tuning on gold data is necessary. First, pre-training data are originally monolingual, so dataset features such as recording quality and volume can be learned instead of linguistic features. Second, the pre-training datasets are from general domains, while the MERLIon dataset contains children-directed speech, which might have a different set of vocabulary.
Nevertheless, with the class imbalance issue, fine-tuning results on MERLIon (exp \#2) improves the BAC but lowers the Mandarin accuracy from 0.650 to 0.630.

\subsubsection{Effect of Gradual Fine-Tuning}
Comparing Experiment \#4 with Experiment \#3, The model's classification accuracy on Mandarin labels improves significantly with a GFT on combined MERLion and SEAME data. 
Despite the class imbalance issue, the  approach (exp \#4) is shown to be successful, allowing the model to effectively extract enough linguistic information from the higher resource but out-of-domain dataset (SEAME) to avoid the short-cut learning from imbalanced in-domain dataset. 

Given the effectiveness of GFT, we further explore experimental designs with the GFT setup combined with data up-sampling to solve the label imbalance issue in the target MERLion dataset. We report the model performance of these additional GFT experiments in Table \ref{tab:gradualresults}. 
First, for the CRNN model, which has a fairly simple residual convolutional neural network architecture, GFT proves to be extremely helpful when fine-tuning on a model pre-trained only on monolingual Mandarin and English data. With a well-design gradual fine-tuning schedule, the classification accuracy on Mandarin improves steadily while the accuracy on English labels is maintained at a reasonable level as shown in Experiment \#14, making this model achieve the best overall results out of all CRNN model variations.

\begin{table}[ht]
\centering
\scalebox{0.82}{\begin{tabular}{cc|ccc}
\hline
\# & CRNN Exp. Desc & English & Mandarin & Balanced \\ \hline
11 & 3ep-GFT lr=1e-3 &  \textbf{0.938}   & 0.270 & 0.604  \\
12 & 3ep-GFT lr=1e-5 &  0.823   & 0.410 & 0.616  \\
13 & 5ep-GFT lr=1e-3 &  0.798   & 0.610 & 0.704  \\
14 & 5ep-GFT lr=1e-5 &  0.932   & \textbf{0.680}  & \textbf{0.806}  \\\hline
\# & MTL Exp. Desc   & \multicolumn{3}{c}{Balanced Accuracy} \\ \hline
15 & non-GFT 20ep     &&  \textbf{0.835}  \\
16 & GFT ep1-5        &&  0.800  \\
17 & GFT ep5-10       &&  0.806 \\
18 & GFT ep10-15      &&  0.817  \\
19 & GFT ep15-20      &&  0.805  \\
\hline
\end{tabular}}
\caption{Performance of the two systems when fine-tuned with up-sampling and gradual fine-tuning.}
\label{tab:gradualresults}\vspace{-2mm}
\end{table}

On the other hand, GFT does not seem to be the contributing factor to the success of the MTL system in predicting the LID of the code-switched utterances. While keeping the MERLion:SEAME data ratio constant, Experiment \#15 achieves the best performance across all systems and designs. This could be explained by the ASR portion of the loss function in the MTL framework, which forces the model to extract higher-level linguistic representations. This increases the robustness of the model against out-of-domain data (SEAME in this case) and therefore the smooth domain adaptation provided by the gradual fine-tuning schedule does not contribute as much in this system design. Up-sampling proves to be extremely helpful in this situation, as Experiment \#15 outperforms Experiment \#7 by 6.4\%. The up-sampling provides the model with more opportunities to learn from and accurately classify instances of the underrepresented class, which leads to a high BAC.

\section{Conclusion}
In this work, we propose two simple and efficient systems for the spoken English-Mandarin child-directed code-switching LID task. The CRNN approach uses a simple stack of CNNs and RNNs to capture information from both the spectral and temporal axes. The multitask learning approach utilizes ASR CTC loss as an auxiliary task to learn higher-level linguistic features for CSLID. Our models significantly outperform previous baselines as well as multilingual PLMs, and we conduct extensive ablation studies to investigate factors that might influence CSLID performance. 
Future works include upsampling the minority label class and fine-tuning PLMs for larger-scale transfer learning to benefit code-switching speech processing. 

\section*{Limitations}
Some of the limitations of our work include the fact that we are not able to use a large batch size when training the model due to compute limits, which might contribute to slower convergence and noisy model performance. Furthermore, we do not leverage cross-lingual transfer from other languages outside of the two languages that are included in the code-switched data. Incorporating code-switched data in other language pairs or monolingual data in related languages might result in additional positive cross-lingual interference.

\bibliography{anthology,custom}
\bibliographystyle{acl_natbib}

\end{document}